\documentclass[10pt, a4paper]{article}

\usepackage{lrec-coling2024} 

\usepackage{amssymb}
\usepackage{amsmath}
\usepackage{fdsymbol}
\usepackage{graphicx}
\usepackage{xcolor}
\usepackage{csquotes}
\usepackage{subcaption}

\usepackage{tikz}

\title{Positive and Risky Message Assessment for Music Products}

\name{Yigeng Zhang\textsuperscript{1}, Mahsa Shafaei\textsuperscript{1}, Fabio A. González\textsuperscript{2}, Thamar Solorio\textsuperscript{1,3}} 

\address{\textsuperscript{1}University of Houston, Houston, USA\\\textsuperscript{2}Universidad Nacional de Colombia, Bogotá, Colombia\\\textsuperscript{3}MBZUAI, Masdar City, United Arab Emirates \\
         \textsuperscript{1}\texttt{\{yzhang168,mshafaei,tsolorio\}@uh.edu}\\
         \textsuperscript{2}\texttt{fagonzalezo@unal.edu.co}}

\abstract{
In this work, we introduce a pioneering research challenge: evaluating positive and potentially harmful messages within music products. 
We initiate by setting a multi-faceted, multi-task benchmark for music content assessment. Subsequently, we introduce an efficient multi-task predictive model fortified with ordinality-enforcement to address this challenge. 
Our findings reveal that the proposed method not only significantly outperforms robust task-specific alternatives but also possesses the capability to assess multiple aspects simultaneously. 
Furthermore, through detailed case studies, where we employed Large Language Models (LLMs) as surrogates for content assessment, we provide valuable insights to inform and guide future research on this topic. The code for dataset creation and model implementation is publicly available at \href{https://github.com/RiTUAL-UH/music-message-assessment}{\texttt{https://github.com/RiTUAL-UH/music-message-assessment}}.
 \\ \newline  }

\begin{document}

\maketitleabstract

\section{Introduction}
Accessing music has never been more convenient than it is today. People can use various tools, such as high-fidelity players and streaming apps, to enjoy music at any time. Listeners can simply go online, press the \emph{PLAY} button, and find themselves invigorated after a bad day.
However, this easy access also raises concerns that children and adolescents may have a higher chance of being exposed to risky content. Young people's thoughts and behavior might potentially be affected by the positive or questionable content in songs, as they tend to learn from the modeled behavior that popular music represents \cite{primack2008content}.
For example, a study has shown an association between adolescent early sexual experiences and degrading sexual content in music \citep{primack2009exposure}. 
Similarly, researchers have also revealed that listening to particular types of songs is positively associated with substance use and aggressive behaviors  \citep{chen2006music}.

The American Academy of Pediatrics (AAP) holds the opinion that parents should be informed of pediatricians' concerns regarding the potentially harmful effects of music lyrics \citep{AAP:96}.
Policymakers have also taken action; for instance, the Recording Industry Association of America (RIAA) introduced the Parental Advisory Label (PAL) program in 1985 to identify audio products with potentially inappropriate content. Such labels were created to draw parents' attention to products that may not be suitable for their children. 
Recent studies from the NLP community have made advances in automating the content rating process. \citeauthor{chin2018explicit} and  \citeauthor{fell2020love} presented machine learning-based methods to automatically classify explicit/non-explicit lyrics in different languages. Further work \cite{info14030159} studied detecting explicitness in several aspects such as strong language and language that refers to violence. The learning objectives are derived from content rating systems like PAL, and these works treat the problem as a binary classification task. Despite these existing achievements, however, neither the current PAL system nor the corresponding automated approaches are capable of appraising a music product with severity information about content suitability. 

In this work, we introduce a novel NLP task: assessing the positive and risky messages of a music item. 
We study the messages that a music item conveys from five significant dimensions regarding appropriateness:  \emph{Positive Messages}, \emph{Violence}, \emph{Substance Consumption}, \emph{Sex}, and \emph{Consumerism} along three degrees of severity. 
In light of those perspectives, we propose a multi-task method that successfully incorporates both content aspect correlations and severity ordinalities to better assess music products.
Our research focuses specifically on music performed with vocal techniques (singing, rapping, etc). 
In such a format of music, the lyrics play a dominant part in conveying the message and opinion from the artists, which has a major impact on the listener’s thoughts and minds. 
Leveraging an automated approach for music content assessment can offer numerous advantages to various stakeholders in the music industry. For music providers, it promises enhanced service quality, particularly benefiting younger listeners and their guardians by facilitating better content recommendations and advisories. Furthermore, lyricists can utilize this system during the early stages of their creative process to assess the portrayal of potentially risky behaviors, such as \emph{`get drunk and be somebody'}. Additionally, the automated identification of questionable content can expedite the creation of clean lyric versions, enabling artists to produce safe renditions more efficiently.

\noindent \textbf{Our contribution:} To the best of our knowledge, this study represents the first exploration into assessing music items, evaluating both their positive and risky dimensions across multiple levels. We have established a comprehensive benchmark and have made the dataset creation method accessible to the wider research community\footnote{\href{https://github.com/RiTUAL-UH/music-message-assessment}{\texttt{https://github.com/RiTUAL-UH/music-message-assessment}}.}. Additionally, we introduce an effective multi-task, ordinality-enforced rating approach that jointly assesses diverse risk factors, delivering state-of-the-art results. As part of our thorough analysis, we also evaluate the efficacy of Large Language Models (LLMs) and provide in-depth discussions, paving the way for future research for this topic.

\section{Music rating and lyrics}
To build up a reliable benchmark for assessing positive and risky messages in such lyrics, we exploit expert ratings provided by Common Sense Media (CSM)\footnote{\href{https://www.commonsensemedia.org}{\texttt{https://www.commonsensemedia.org}}.}. CSM is a non-profit organization caring for kids' digital well-being. It hosts a media rating platform that is supported by childhood development experts. The rating system covers various aspects of childhood development and age appropriateness. 
In this work, based on the focal points from Youth Risk Behavior Surveillance System (YRBSS) \cite{kolbe1993overview} introduced by the USA Centers for Disease Control and Prevention, we choose three explicitly risky aspects \emph{Violence}, \emph{Substance Consumption}, and \emph{Sex}. We also pick \emph{Consumerism} that relates to drawing interest in the acquisition of goods, often associated with a ``feel good" experience. Because \emph{Consumerism} potentially affects the emotional health and identity development in youth \cite{hill2011endangered}. Our dataset also includes a category dedicated to \emph{Positive Messages} within music products. The automatic detection of such uplifting content offers numerous valuable applications, enhancing user experience. This could serve as a counter to the often conflicting messages found in mainstream lyrics. Furthermore, it might be leveraged as a form of supportive therapy to positively influence users' moods.

\subsection{Dataset facts and analysis}
With CSM expert ratings, we developed a dataset based on publicly available lyrics from the Internet. We collected 1,119 music items (consisting of 10,661 songs) including standard albums, extended plays (EP), long plays (LP), and CD singles. We refer to the collections of songs as \emph{albums} in this work for simplification. The partition of CD singles and the albums are shown in Table \ref{tab:albums_percentage}. The \emph{age recommendation}  spans from 2 to 18+. Notably, the music items we consider in this work are in English and comprise a heavily Western-centric dataset.

\begin{table}[h]
\centering
\resizebox{0.7\columnwidth}{!}{%
\begin{tabular}{ccc}
\hline
Category & Album & CD Single \\
\hline
Percentage & 62.2\% & 37.8\% \\
\hline
\end{tabular}
}
\caption{Percentage of albums and CD singles.}
\label{tab:albums_percentage}
\end{table}

\begin{figure}[htbp]
    \centering
    \begin{subfigure}{0.49\columnwidth}
        \includegraphics[width=\columnwidth]{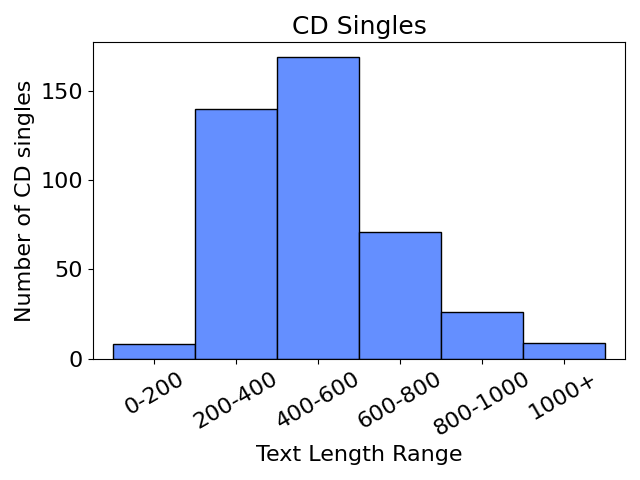}
    \end{subfigure}
    \begin{subfigure}{0.49\columnwidth}
        \includegraphics[width=\columnwidth]{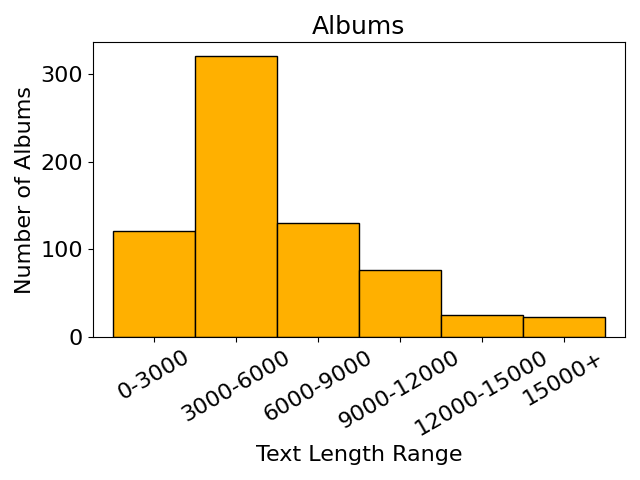}
    \end{subfigure}
    \caption{Lyrics length distribution for CD singles and albums.}
    \label{fig:text-length}
\end{figure}

The duration of individual music items, including both CD singles and albums, shows significant variation. Comprehensive distributions of text lengths are illustrated in Figure \ref{fig:text-length}.

The level of aspect prevalence is from 0 to 5. To reduce class imbalance due to the fact of data availability, we then project such scored ratings into 3-level ordinal categories (Low Presence (0-1), Medium Presence (2-3), High Presence (4-5)) with median split strategy as in related works \citep{martinez2019violence, martinez-etal-2020-joint} from the movie domain.
A detailed label distribution is described in Table \ref{tab:number-aspect}.
\begin{table*}[h!]
\centering
\resizebox{0.6\linewidth}{!}{%
\begin{tabular}{l|ccccc}
\hline
Aspect & Violence & Substance & Sex & Consumerism & Positive \\ \hline
Low  & 844   & 743     & 663      & 880    & 736        \\
Medium    & 190    & 294      & 319       & 209     & 314         \\
High   & 85    & 82      & 137       & 30     & 69         \\ \hline
\end{tabular}%
}
\caption{Low/medium/high presence item distribution for each message aspect.}
\label{tab:number-aspect}
\end{table*}


The music rating data comes from Common Sense Media (CSM). Non-member users can only browse up to three expert product reviews for free each month. We gained permission from CSM to use the music rating data for research, however, by the time we submit this work, the music reviews are no longer shown on CSM's website until further updates. We still release the names of expert-rated products and links to their lyrics studied in this work for the community. We do not release the lyrics directly due to copyright considerations; however, lyrics can be easily accessed through the links we provide and via search engines. While we've made every effort to collect lyrics from the open Internet, we acknowledge that some songs might be missing from certain albums.
\subsubsection{Inter-dimensional correlation analysis}
\begin{figure}[ht]
\centering
  \includegraphics[width=0.9\linewidth]{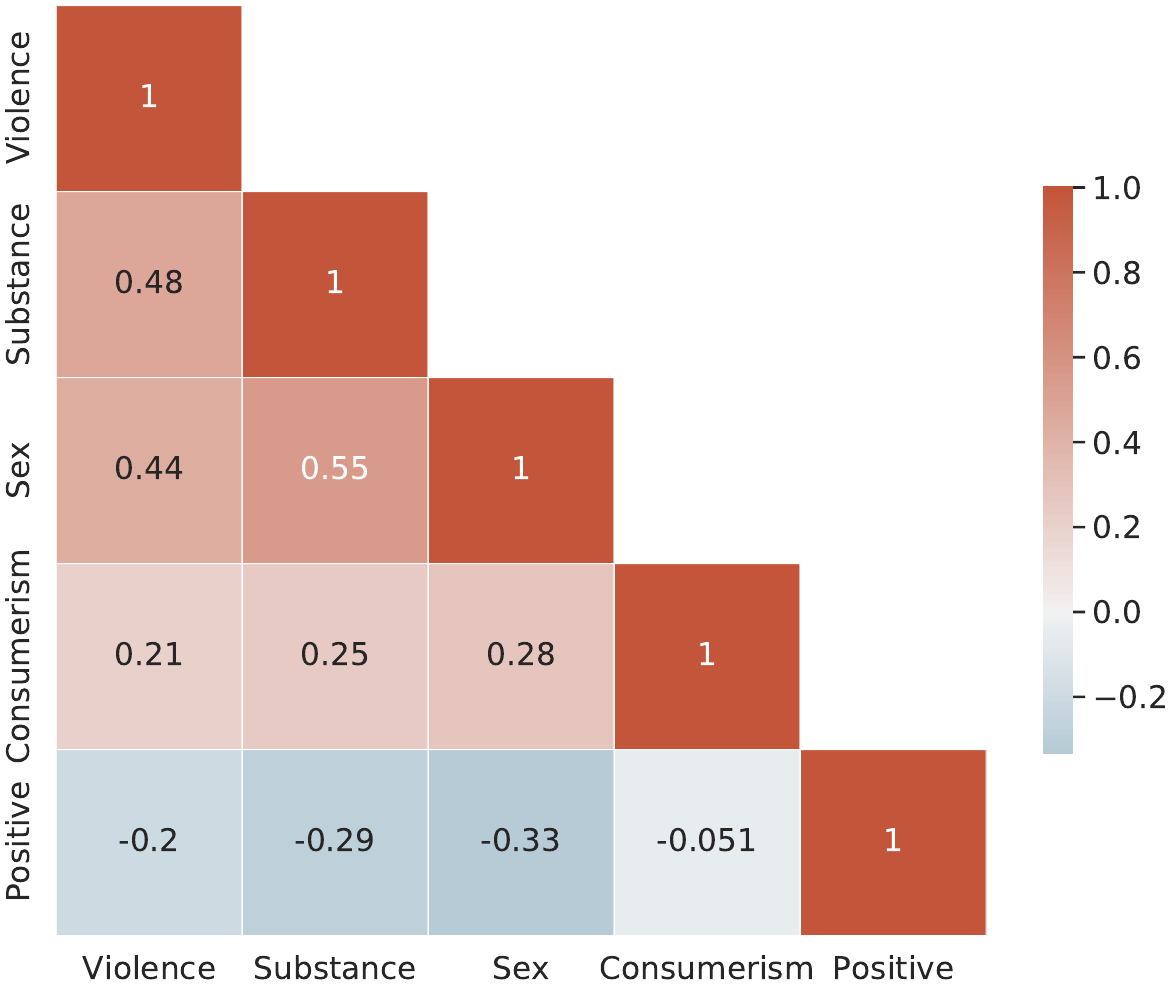}
  \caption{Spearman rank correlation between positive and risky prevalence pairs.}
  \label{fig:spearman}
\end{figure}

Empirically, risky behaviors such as violence and substance use often appear concurrently in a piece of music. We further calculate the Spearman rank correlation $\rho$ between each positive and risky rating pair among all of the music items. The correlation heat map is demonstrated in Figure \ref{fig:spearman}. All correlation scores between variable pairs have significance with $p < .05$ except \emph{Positive Messages-Consumerism}.

We can easily observe from the heat map that typical physical risky behaviors  \emph{Violence},  \emph{Substance Consumption}, and \emph{Sex} have a positive correlation with each other, while \emph{Consumerism} has a positive correlation with those three but less strong. It is no surprise that \emph{Positive Messages} has a significant negative correlation with three physical risky behaviors. 
This interdimensional behavior intuitively inspires us to leverage such correlations to design relevant machine learning strategies. 

\subsubsection{Positive/Risky behavior \emph{v.s.} Explicitness}

\begin{figure}[ht]
\centering
  \includegraphics[width=0.99\linewidth]{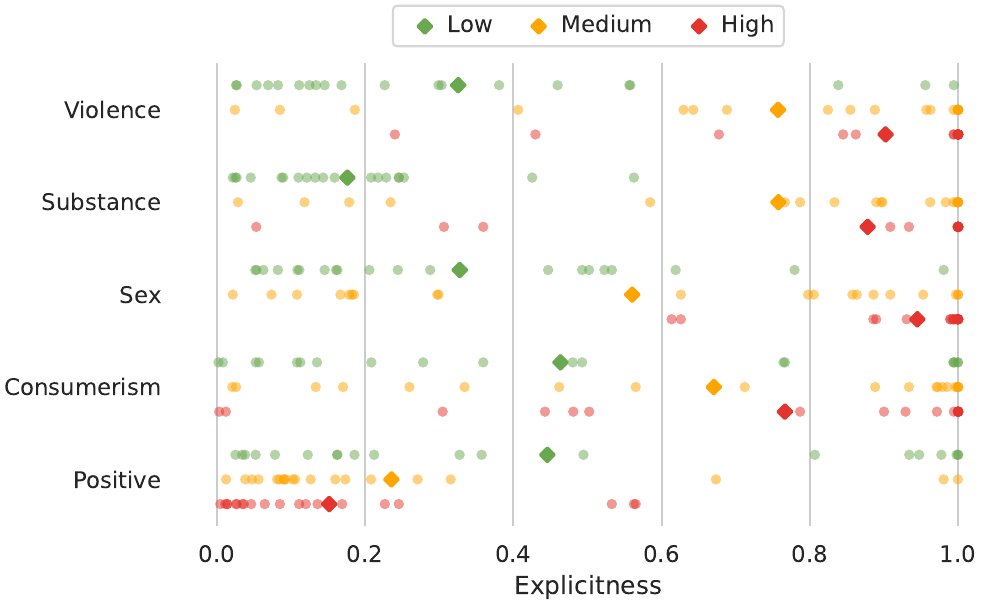}
  \caption{Explicitness probability on different message dimensions. 20 instances are randomly sampled from each level. The diamond symbol $\medblackdiamond$ indicates the central tendency of a series of probability values for the corresponding level.}
  \label{fig:explicit}
\end{figure}

In the context of music, \emph{explicitness}  often refers to content that contains strong language or violence, sex, or substance abuse depictions (from RIAA). It is a generalized description that has a high coincidence with the risky message we studied in this work. Since it is not practical to collect gold labels of explicitness for every music item we studied, we apply an explicit lyrics classifier trained on 438k English lyrics from a previous work \cite{fell2020love}. We can further explore the correlation between the explicitness probability and the level of positive and risky messages. Figure \ref{fig:explicit} illustrates the pattern of explicitness probability towards severity levels of different aspects.

The message ratings consistently correlate with explicitness values. For all risky message aspects, explicitness increases as ratings transition from low to high. Even the abstract message of \emph{Consumerism} aligns with this trend. In contrast, \emph{Positive Messages} exhibit a distinct negative correlation with explicitness. These correlations suggest that message ratings can serve as a valuable complement or alternative to traditional metrics of music appropriateness. They provide a richer perspective and highlight the significance of our study.

\begin{figure}[h!]
\centering
  \includegraphics[width=0.95\linewidth]{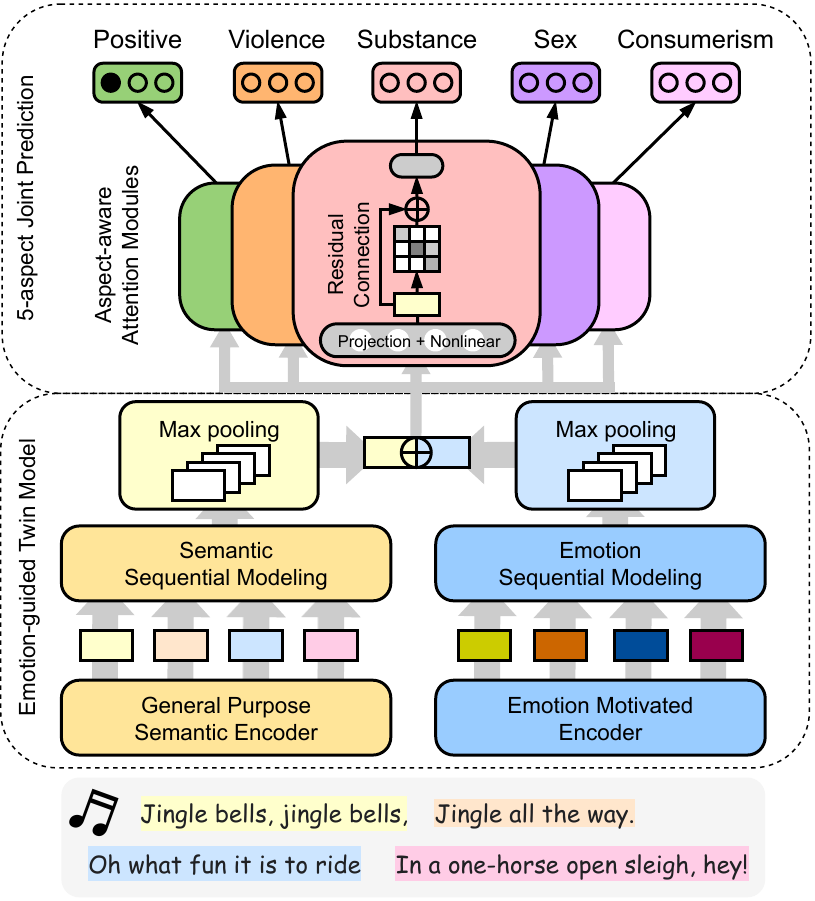}
  \caption{Joint prediction architecture with emotion-guided twin and aspect-aware attention module.}
  \label{fig:arch}
\end{figure}

\section{Methodology}
We formulate this music content assessment problem as a multi-class text classification task, where the ratings of an aspect are the prediction objectives. We take solely lyrics as visible features. 
Expecting to leverage the correlation pattern among different messages and their ordinal levels, we propose an effective model that incorporates rich semantic representation, aspect-aware multi-task learning, and ordinality-enforcement. The model architecture is illustrated in Figure \ref{fig:arch}.

\textbf{Emotion-guided twin model:} We propose to encode the text from two perspectives: general-purpose semantic representation and emotion-centered semantic representation, as emotion information has been proved to be an effective complement for various language understanding tasks \cite{shafaei-etal-2020-age, samghabadi2020attending}. We employ two pretrained Transformer models finetuned with semantic textual similarity (Sentence-BERT \cite{reimers-gurevych-2019-sentence}) and emotion detection (task-specific Distilled RoBERTa on emotion detection \cite{hartmann2022emotionenglish}) respectively. The final representation is the concatenation of these two types of embeddings.

\textbf{Joint prediction over multiple aspects:}
From the dataset study, we found different types of positive and risky messages, specifically questionable behaviors, have a potential correlation with each other. For instance, the presence of violence in a music item might appear concurrently with lyrics depicting substance use. The learning objective is a joint loss of predicting ratings of multiple aspects. 
To fortify the representation uniqueness for different aspects, we further design an aspect-aware attention module to learn specialized features for each aspect. The module begins with a projection layer with non-linearity and further learns weights through an aspect-differentiation matrix. The final representation $\boldsymbol {x}_{out}$ comes from the addition of input $\boldsymbol {x}_{in}$ with residual over skip connections as described in Equation \ref{eq:aamodule}. 
\begin{equation}\label{eq:aamodule}
\boldsymbol{x}_{out}= \boldsymbol{x}_{in} +  \boldsymbol{x}_{in} \circ {Softmax}\left(\boldsymbol{x}_{in} {\boldsymbol{W}}_{attn}\right)
\end{equation}

\textbf{Ordinality-enforcement techniques:} 
The severity ratings are discrete interclass ordinal variables instead of independent categorical labels. Typical classification models usually ignore such correlations between ordinal categories. In this work, we apply three ordinality-enforcement techniques that are better suited for the task. All of the ordinality-enforcement techniques are applied respectively to the base model with a multi-tasking module. 
\begin{itemize}
    \item \textbf{\emph{Siamese ranking-classification}} \cite{zhang-etal-2021-none-severe}: This method leverages a Siamese network to process a pair of instances for both ranking and classification objectives. The ranking (comparison) operation has the potential to learn pairwise ordinal differences in severity levels between samples. The model is optimized with two cross-entropy losses for the two objectives as in Equation \ref{eq:rank}, where $l_{\text{cls}}$ comes from multi-class classification while $l_{\text{rank}}$ is derived from comparing ratings (lower/same/higher) between the two music items.
    \begin{equation}\label{eq:rank}
    \hat{f} \leftarrow \underset{f}{\arg \min }\left(l_{\text{cls}}+l_{\text{rank}}\right)
    \end{equation}
    \item \textbf{\emph{Binary attributes transformation}} \cite{frank2001simple}: 
    This method tackles ordinal regression by dividing the sorted ordinal label set, containing $n$ elements, into two subsets at every possible pair of adjacent elements. This results in $n-1$ potential splits. Each split transforms the ordinal regression problem into a binary classification task, where the goal is to predict whether the ordinal value $y_{i}$ falls before or after the split point within the set. 
    This approach applies multiple classifiers to leverage the ordinal information repeatedly. Specifically, in the setting of this problem with three ordinal classes (low, medium, and high), we consider two binary splits: one between low and medium, and another between medium and high. This results in two binary classifications:
    \begin{equation}
    \begin{aligned}
    \operatorname{Pr}(y_{i} \leqslant \text{Low}), \operatorname{Pr}(y_{i} > \text{Low});
    \\
    \operatorname{Pr}(y_{i} \leqslant \text{Mid}), \operatorname{Pr}(y_{i} > \text{Mid}).
    \end{aligned}
    \end{equation}
    In this setting, for each classification, we apply a binary classifier to predict which split the prediction will fall in, utilizing the ordinal information multiple times aiming to improve the prediction performance.
    \item \textbf{\emph{Soft label}} \cite{diaz2019soft}: This method introduces a label softening function to convert ordinal category values into a probability distribution across categories. A class label $y_i$ from the label set is encoded into a soft label $c_{i}^{soft}$ using the following formula for a specific true rating rank $y_{t}$. 
    \begin{equation}
    c_{i}^{soft}=\frac{e^{-\phi\left(y_{t}, y_{i}\right)}}{\sum_{k=1}^{K} e^{-\phi\left(y_{t}, y_{k}\right)}} \quad \forall y_{i} \in \mathbb{Y}.
    \end{equation}
    The Kullback-Leibler divergence is used as the loss function to measure the difference between the predicted probability and the soft label.
    $\phi\left(y_{t}, y_{i}\right)=\left|y_{t}-y_{i}\right|$ is chosen as the metric penalty for the sake of simplicity.
\end{itemize}

\noindent

\begin{table*}[h!]
\centering
\resizebox{0.95\textwidth}{!}{%
\begin{tabular}{lcccccc}
\hline
\multicolumn{1}{l|}{\textbf{Baselines}}                  & Violence       & Substance      & Sex            & Consumerism    & \multicolumn{1}{l|}{Positive}       & Avg            \\ \hline
\multicolumn{1}{l|}{Majority voting}                     & 28.65          & 26.59          & 24.76          & 30.22          & \multicolumn{1}{l|}{26.46}          & 27.34          \\
\multicolumn{1}{l|}{TF-IDF + SVM}                 & 51.28          & 47.31          & 61.16          & 40.51          & \multicolumn{1}{l|}{27.79}          & 45.61          \\
\multicolumn{1}{l|}{BoWV + SVM}                   & 46.75          & 40.23          & 55.63          & 31.15          & \multicolumn{1}{l|}{28.28}          & 40.41          \\
\multicolumn{1}{l|}{TextCNN \cite{kim-2014-convolutional}}                      & 50.59          & 48.25          & 59.03          & 44.88          & \multicolumn{1}{l|}{36.79}          & 47.91          \\
\multicolumn{1}{l|}{TextRCNN \cite{Lai_Xu_Liu_Zhao_2015}}                     & 57.63          & 55.86          & 64.01          & 44.61          & \multicolumn{1}{l|}{38.47}          & 52.12          \\ \hline	
\multicolumn{1}{l|}{LSTM+Attention \cite{shafaei-etal-2020-age}}               & 33.93          & 34.48          & 41.65          & 34.62          & \multicolumn{1}{l|}{30.06}          & 34.95          \\ 
\multicolumn{1}{l|}{BERT \cite{devlin-etal-2019-bert,fell2020love}}                         & 56.93          & 52.16          & 60.61          & 44.96          & \multicolumn{1}{l|}{\textbf{46.57}}          & 52.25          \\
\multicolumn{1}{l|}{RNN Trans (RT) \cite{zhang-etal-2021-none-severe}}               & 62.84          & 62.47          & 67.36          & 46.16         & \multicolumn{1}{l|}{44.37}          & 56.64          \\ \hline

\multicolumn{7}{l}{\textbf{multi-task + ordinality-enforcement}}                                                                                                                                   \\ \hline
\multicolumn{1}{l|}{Soft label}       & 61.42  & 64.36         & 68.92          & 45.21          & \multicolumn{1}{l|}{42.60}          & 56.50          \\
\multicolumn{1}{l|}{Ranking-classification}    & \textbf{65.04} & \textbf{64.41}          & \textbf{69.11} & 45.65          & \multicolumn{1}{l|}{44.72}          & 57.79          \\
\multicolumn{1}{l|}{Binary transformation}     & 64.46          & 63.98          & 69.00          & \textbf{47.36}          & \multicolumn{1}{l|}{44.61}          & \textbf{57.88} \\ \hline
\end{tabular}%
}
\caption{Experimental result on positive and risky message level with macro F1 scores with 10-fold cross-validation.}
\label{tab:exp}
\end{table*}

\section{Experiments and Results}


To evaluate the effectiveness of our proposed method in music content assessment, we benchmarked several popularly used classification methods and models from related works in media rating. We choose TF-IDF and Bag-of-Word-Vectors (Averaged GloVe \cite{pennington-etal-2014-glove} embeddings) with SVM classifiers,  TextCNN \cite{kim-2014-convolutional} and TextRCNN \cite{Lai_Xu_Liu_Zhao_2015}. We experiment with three deep models that are designed for media rating problems: 
\begin{enumerate}
    \item An RNN model with attention for predicting movie MPAA ratings based on movie dialogue scripts \cite{shafaei-etal-2020-age} as lyrics are also sequences of utterances;
    \item A BERT model \cite{devlin-etal-2019-bert}  as used for classifying explicitness in \cite{fell2020love};
    \item An RNN-Transformer backbone model of state-of-the-art in rating severity for age-restricted content in movies \cite{zhang-etal-2021-none-severe}.
\end{enumerate}
The training-development-test split uses an 80/10/10 ratio with data shuffled. We choose macro F1 as the classification performance metric because the label distribution from the dataset is highly imbalanced. Experimental results are shown in Table \ref{tab:exp}.

Among deep baseline models, there is no dominant architecture that can give the best prediction performance on every aspect but the RNN-Transformer (RT) model has an overall best performance with a notable gap. Our proposed method, emotion-guided multi-tasking model with ordinality-enforcement, shows an overall best performance among all methods. The repeated-measures t-test shows our ordinality-enforcement models give a significant performance improvement ($p < .05$) over the strongest baseline on average, specifically on explicit risky behaviors (\emph{Violence}, \emph{Substance consumption}, and \emph{Sex}), using five random seeds with 10-fold cross-validation mean F1 score.

\section{Discussion and analysis}

Speaking of aspects, three types of explicit content, \emph{Violence}, \emph{Substance}, and \emph{Sex} are relatively easier for the model to predict than \emph{Consumerism} and \emph{Positive messages}. The reason could be that the latter aspects are more abstract concepts compared to risky behaviors that can be described in the lyrics in an overt manner. For \emph{Consumerism}, we hypothesize that a model which can more effectively capture signals related to 'goods' and the promotion of purchases might perform better. Unlike explicit questionable content like \emph{Violence}, \emph{Positive messages} is even harder to intuitively find a clear textual pattern since \emph{positive} is more heterogeneous than the other aspects. We may need the model to gain a high-level understanding of the content to distinguish the quantity of positive value that a song or an album delivers.

\subsection{Ablation study}

We conducted an ablation study on the best-performing model by iteratively removing individual components from the model architecture. The experiment demonstrated that all modules within the network structure contributed to the performance in prediction. When we configured the proposed network for single-task prediction with all modules, the performance reached even higher levels. This indicates that the effectiveness of the novel network modules can help single-task baseline models learn better quality text representations. Additionally, we experimented with the backbone model by having it predict all five aspects directly in a multitask setting, which resulted in a large performance drop. This finding highlights the effectiveness and necessity of incorporating components that can guide and enhance multitask learning.

\begin{figure*}[ht]
\centering
  \includegraphics[width=0.9\linewidth]{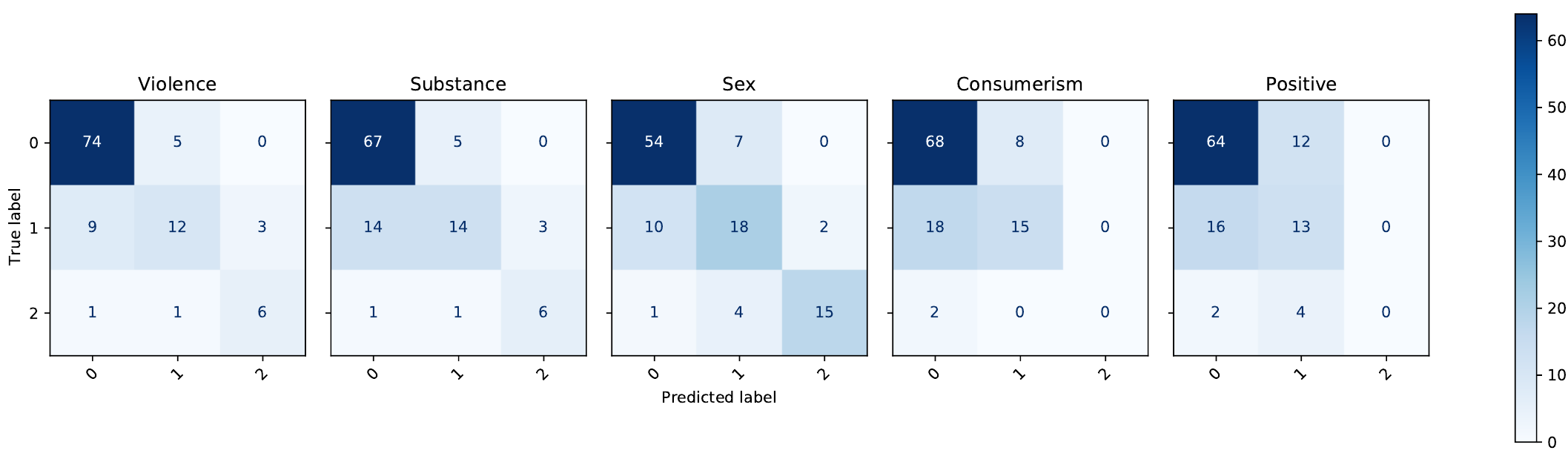}
  \caption{Prediction confusion matrix of the best-performing method on all 5 aspects. The x-axis indicates the predicted values and the y-axis indicates the ground-truth labels.}
  \label{fig:confmat}
\end{figure*}

\begin{table}[h]
\centering
\resizebox{0.9\columnwidth}{!}{%
\begin{tabular}{l|c}
\hline
Ablation                      & Perf change \\ \hline
Best performing model         & 57.88               \\ \hline
Aspect-aware module           & 57.58 (-0.30)              \\
Emotion-guided twin model     & 57.77 (-0.11)              \\
Ordinality enforcement (binary) & 57.53 (-0.24)              \\
Multitask joint prediction    & 59.05 (1.17)               \\ \hline
Backbone only + multitask     & 55.36 (-2.52)              \\ \hline
\end{tabular}%
}
\caption{Ablation study of different components in the best-performing model. We report and analyze the average performance changes across the five aspects.}
\label{tab:ablation}
\end{table}

\begin{table*}[h!]
\centering
\resizebox{0.92\textwidth}{!}{%
\begin{tabular}{llccccc}
\hline
  & Removed sentence from a segment of \emph{Heartless (2019)}                                                      & Violence & Substance & Sex \\ \hline
  & Rating                                                       & Low        & High         & Mid          \\
  & Whole segment confidence                                     & 98.77    & 90.57     & 72.60   \\ \hline
1 & Stix drunk, but he never miss a target                       & 1.13     & -90.56 $\Downarrow$    & -64.57 $\downarrow$      \\
2 & Photoshoots, I'm a star now (Star)                           & -4.18   & 2.48      & -40.27 $\uparrow$      \\
3 & I'm talkin' Time, Rolling Stone, and Bazaar now (Bazaar now) & -1.91    & 1.55      & -15.39      \\
4 & Sellin' dreams to these girls with their guard down (What?)  & -1.71    & 2.01      & -30.90 $\downarrow$        \\ \hline
\end{tabular}%
}
\caption{An input perturbation study on the behavior of the proposed ranking-classification model. We choose three risky aspects - \emph{Violence}, \emph{Substance consumption}, and \emph{Sex} - as this model yields the best performance. The numbers indicate the absolute probability change of the original prediction result. A double down arrow $\Downarrow$ indicates the predicted severity downgraded by two levels, a single down arrow $\downarrow$ means downgraded by one, and an up arrow $\uparrow$ represents upgraded by one.}
\label{tab:perturbation}
\end{table*}

\subsection{Saliency analysis}
We perform saliency analysis using input perturbation to better understand the model prediction behavior. We chose a segment from the lyrics of \emph{Heartless (2019)} by \emph{The Weekend}, which contains explicit language. We do the perturbation sentence by sentence through removing one and feeding the rest of the lyrics into the model, then we inspect the model prediction result. Table \ref{tab:perturbation} shows the detailed influence on prediction confidence and outcome. 
For \emph{Substance Consumption}, line 1 explicitly contains the word \emph{drunk}. When we remove this sentence, the prediction drastically changes from \emph{High} to \emph{Low}.   
For \emph{Sex}, when we remove the first line, the prediction probability decreases and the result becomes \emph{Low}. The same situation happens to the fourth line. Interestingly, removing line 2 results in an upgrade. We hypothesize that this deletion increases the density of sexual implication content.
This case study intuitively shows the model can successfully capture particular mentions of risky messages such as substance use and sex-related topics that have significance in severity prediction.

\subsection{Error analysis}
Figure \ref{fig:confmat} shows the confusion matrix of the prediction results of the best-performing model from a single fold of the cross-validation. In general, the proposed method can capture the ordinal information well because wrong predictions that cross two levels (predict low to high or high to low) are rare. Specifically, the model struggled to give correct predictions on high \emph{Consumerism}. We suppose the number of training instances of high \emph{Consumerism} is relatively small. The same case happened to high \emph{Positive} predictions. We hypothesize the heterogeneous nature of \emph{Positive} content makes it challenging to predict. 

\begin{figure*}[ht]
\centering
  \includegraphics[width=0.9\linewidth]{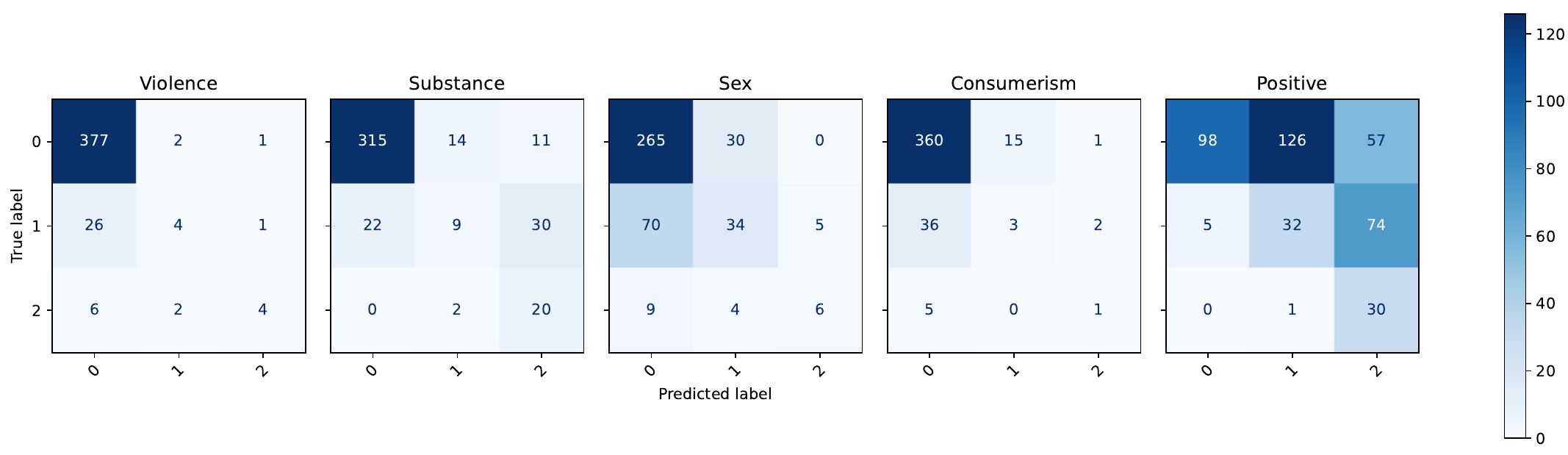}
  \caption{Prediction confusion matrix of the best-performing LLM method on all 5 aspects. The x-axis indicates the predicted values and the y-axis indicates the ground-truth labels.}
  \label{fig:chatgpt_confmat}
\end{figure*}

\subsection{Case study: unsuccessful predictions}
We dig into some unsuccessful predictions to analyze the errors. We mainly focus on the hard aspects of the model.
\begin{itemize}
    \item \textbf{Album: The Best Damn Thing (2007) by Avril Lavigne}: The proposed model gives \emph{Consumerism} low and \emph{Positive} low ratings, however, the correct labels for those two aspects are medium. This album is a pop-punk production and the songs in the album seemed to be targeted at young people with themes such as love and encouragement. For \emph{Consumerism}, there are explicit lyrics saying: 
    \begin{displayquote}
    \emph{I hate it when a guy doesn't get the tab}
    
    \emph{And I have to pull my money out, and that looks bad}
    \end{displayquote}
    But such cases are rare and one will not make a confident decision for a strict medium rating. For \emph{Positive}, this album contains lyrics with significant positive values such as \emph{Keep Holding On}:
    \begin{displayquote}
    \emph{You're not alone}
    
    \emph{Together we stand}
    
    \emph{I'll be by your side, you know I'll take your hand}
    \end{displayquote}
    We suspect that explicit expressions of positivity as in the example are sparse in the songs. However, many songs in the album convey a spirit of pursuing happiness and love. This nuance gives the album a medium rating in \emph{Positive}, but it is challenging for the model to capture.
    \item \textbf{Album: A Hard Day’s Night (1964) by The Beatles}: The model gives a critical wrong prediction on \emph{Positive}: predict as low while the ground truth is high. The majority of the songs in this album express deep and genuine emotions about love, with lyrics like:
    \begin{displayquote}
    \emph{If you need somebody to love, just look into my eyes}
    
    \emph{I'll be there to make you feel right}
    
    \emph{If you're feeling sorry and sad, I'd really sympathize}
    
    \emph{Don't you be sad, just call me tonight}
    \end{displayquote} 
    One possible reason the model failed is that the lyrics do not contain words that explicitly convey strong positive signals of companionship. However, the overall sentiment is clearly positive and constructive.
    
    \item \textbf{CD single: Labels or Love (2008) by Fergie}: The model gives \emph{Sex} medium, \emph{Substance consumption} medium, and \emph{Consumerism} low, while the correct answer are \emph{Sex} low, \emph{Substance consumption} low, and \emph{Consumerism} high. For \emph{Sex} medium and \emph{Substance consumption}, the false positive may come from lexical signals such as \emph{sexy}, \emph{kiss} and \emph{bag}.
    For \emph{Consumerism}, there are not only expressions to the encouragement of buying goods:
    \begin{displayquote}
    \emph{Let's stop chasing those boys and shop some more}
    \end{displayquote} 
    but also many explicit mentions of luxury brands:
    \begin{displayquote}
    \emph{Gucci, Fendi, Prada purses, purchasing them finer things}
    
    \emph{Men, they come a dime a dozen}
    
    \emph{Just give me them diamond rings}
    
    \emph{I'm into a lot of bling, Cadillac, Chanel and Coach}
    \end{displayquote} 
    We suspect that the model did not interpret a direct narrative about shopping as a strong indicator of \emph{Consumerism}. Additionally, the names of luxury brands might be rare in the corpus, leading to a lack of supervision signals. As a result, the model struggled with this prediction.

\end{itemize}

\begin{table*}[h]
\centering
\resizebox{0.65\linewidth}{!}{%
\begin{tabular}{l|ccccc|c}
\hline
                  & Vio            & Sub            & Sex            & Con            & Pos            & Avg            \\ \hline
Simple context            & 54.91          & 49.87          & 49.09          & 45.35          & 34.35          & 46.71          \\
Rich context       & 53.51          & \textbf{54.06} & 53.79          & 40.94          & \textbf{35.33} & \textbf{47.53} \\
Simple context + CoT      & 53.72          & 37.83          & 43.22          & \textbf{45.95} & 29.77          & 42.10          \\
Rich context + CoT & \textbf{55.52} & 50.18          & \textbf{54.03} & 40.82          & 29.05          & 45.92          \\ \hline
\end{tabular}%
}
\caption{Zero-shot evaluation results from \texttt{gpt-3.5-turbo}. Only CD-singles are evaluated due to the intrinsic token size limitations of this LLM.}
\label{tab:chatgpt}
\end{table*}

\subsection{Case study: LLM as content judges}
Recent advancements in Large Language Models (LLMs) have showcased impressive natural language understanding and adaptability across a multitude of tasks. Motivated by these advancements, our study aims to explore the potential of LLMs in assessing content within music products. Specifically, we leverage the \texttt{gpt-3.5-turbo} API \cite{OpenAI2022, ouyang2022training} as surrogate evaluators. Our primary focus is on rating five specific aspects of the content. The central hypothesis of this study is that, despite their inherent limitations and lack of access to a supervision signal, LLMs can provide content assessments that are both meaningful and comparable in accuracy to other deep learning methods. Due to the context length constraints of LLMs, we limit our evaluation to CD-singles, setting a token cap of 3000 to ensure the model's efficient functioning.

The experiments were structured in three distinct formats:
\begin{itemize}
    \item Simple Context: The LLM is directed to rate each song across the five aspects without any supplementary information.
    \item Rich Context: Before prompting the LLM, a detailed description of the five aspects is provided in the context.
    \item Chain-of-Thought (CoT): Building upon the CoT approach \cite{wei2022chain}, known for enhancing LLM performance in complex reasoning tasks, we feed the model with exemplar prompts as context and subsequently instruct it to complete the rating task.
\end{itemize}

Table \ref{tab:chatgpt} presents the evaluation results. While LLM-based approaches have their merits, they did not demonstrate remarkable efficiency in this content rating context. Although providing a richer context yielded marginally superior outcomes, no method consistently outperformed the others. Interestingly, LLM assessments aligned with patterns observed in our baseline and proposed methods. Specifically, the LLM found it more straightforward to evaluate the explicit aspects of \emph{Violence}, \emph{Substance}, and \emph{Sex}, but faced challenges with \emph{Consumerism} and \emph{Positive}. It's important to note that this experimental design is simpler than previous sections, given it exclusively assesses CD-singles rather than an assorted selection from an album.

Further analysis of the LLM's performance was conducted by examining the confusion matrix of the top-performing model, as depicted in Figure \ref{fig:chatgpt_confmat}. The model displayed some notable patterns in its behavior. Specifically:
\begin{itemize}
    \item For categories like \emph{Violence}, \emph{Sex}, and \emph{Consumerism}, the model tended to underestimate their respective severities.
    \item When rating \emph{Substance}, the model frequently struggled to give \emph{medium presence} ratings.
    \item In contrast, the evaluation of \emph{Positive} content often resulted in an overestimation of a song's positive messages, different from patterns observed in prior experiments of baseline and proposed methods.
\end{itemize}

It's essential to recognize the intrinsic limitations of the LLM. We could not apply the same training and assessment methods to the LLM as we did in previous sections. Although CD singles are a subset of the broader music product collection, they retain a consistent data property. While our comparison does not strictly align with traditional comparative analysis standards, due to potential disparities in data distribution and features across datasets, it nonetheless provides valuable insights. These insights can guide model benchmarking and optimization, even if not strictly empirical. Our decision to evaluate closed models like \texttt{gpt-3.5-turbo}—which is non-reproducible due to its proprietary nature—stems from a desire to explore the capabilities of such models. We urge readers to interpret these particular results as exploratory, rather than as fixed benchmarks.

While LLMs have demonstrated proficiency in a variety of NLP tasks, their performance in our specialized context of content assessment was not on par. This discrepancy is understandable given that these versatile models are not trained for such tasks. Consequently, their judgments might not always align with the expert opinions of professionals in media research and childhood development. Recognizing this, our forthcoming research aims to explore the development and analysis of task-specific LLMs for content safety. We are optimistic that such an approach will yield meaningful insights and enhanced performance.

\section{Conclusion}

In this paper, we introduce a novel task to the NLP community: predicting the intensity of various aspects of music, spanning from objectionable content to positive messages. By analyzing music product lyrics, we investigate multiple dimensions of messages conveyed to listeners. Our research problem and approach are intended to foster deeper investigations into music content assessment. The multi-task ordinality-enforcement model we present has shown promising effectiveness for this type of challenge with ordinal properties. The case studies, along with our exploration using Large Language Models (LLMs) as surrogate evaluators, highlight the inherent complexities of the message assessment problem, calling for the need for continued community engagement and research.

\section*{Ethical considerations and limitations}
This work introduces a novel task: assessing positive and risky messages for music products. It also proposes a state-of-the-art method to automatically accomplish the assessment. We acknowledge the potential limitations and ethical considerations by highlighting the following points for future explorations on similar topics:

\textbf{Reliability:} We recognize the potential issues regarding the reliability of such a content rating system. Possible inaccuracies may result in misleading content suggestions, potentially leading vulnerable groups to inadvertently consume inappropriate content, or causing confusion in the production processes for musicians. This work represents our initial exploration, and we strongly advise against implementing such a system in real-world services until the technical and operational elements can be held accountable. We insist that such a system should be regarded as an assistant to, rather than a replacement for, the content rating and assessment work done by media experts and customers.

\textbf{Social context concerns:} The social context in which these labels are acquired is not always known, and there can be a lack of context in terms of how language is used and judged. For instance, many rap songs discuss the harms associated with isolation and substance abuse, yet such information might be misclassified due to rating provider bias or system bias. This may increase the likelihood of systematic bias or unintentionally promote racism. Future research should explore distinguishing between racist and reclaimed uses of slurs as well as between mentions of risky subjects in a suggestive manner and those deemed more innocuous.

\textbf{Data source concerns:} Our rating data comes from CSM, a non-neutral organization. The regulations they use to recruit human experts to rate media products are unknown, and different experts rate different products, which may lead to inconsistencies. Transparency and accountability are not guaranteed, and subjectivity remains in the rating results. Future research and implementations should not rashly take ratings from data sources such as CSM as golden standards without careful assessment.

\textbf{Ambiguity in aspects:} The rating aspects used in this work can be ambiguous, as they are loosely defined by single words. For example, what is considered \emph{violence} may extend beyond overtly violent behaviors. We also recognize that the \emph{positive} aspect defined by the CSM, indicating the overall takeaway, lacks fine-grained elaborations compared to risky aspects. This could result in ambiguity and fail to provide further insights for users.

\textbf{Implications and acceptable use:} An ethical concern of this study is the potential for media censorship. We acknowledge that efficient machine learning-based algorithms like the proposed method could be used as censorship tools. Malicious users could misuse the proposed method for illegitimate censorship, potentially harming freedom of speech. We call for the development and use of AI algorithms with special attention to who should use the system, how it should be used, and what safeguards should be in place to prevent misuse.

\textbf{Modality coverage:} A technical limitation of this work is that it does not take other modalities of music products, such as melody, rhythm, and vocal performance, into account for predictions. We recognize the significance of these signals in conveying a song's message. Additionally, we focused only on songs in English in this study, which means lyrics written in other languages and from different cultural backgrounds are absent from our study. Our work stands as an initial exploration of the feasibility of solving this task. We hope that future work can explore diversifying the dataset and exploring how the models behave in those cases.

\section*{Acknowledgements}
We thank Common Sense Media for permitting us to use the expert ratings for research. We would like to thank the anonymous LREC-COLING reviewers for their feedback on this work.

\nocite{*}
\section{Bibliographical References}\label{sec:reference}

\bibliographystyle{lrec-coling2024-natbib}
\bibliography{lrec-coling2024-example}


\appendix

\section{Implementation details}
We list the implementation details of the proposed baseline methods.

\textbf{Sparse and dense document representation:} Here we apply TF-IDF and Bag-of-Word-Vectors (GloVe \cite{pennington-etal-2014-glove} Average) as text vectorization approaches and linear models as baselines. TF-IDF method is also used in the previous work that classifies explicitness in lyrics \cite{fell2020love}.

\textbf{Word-level semantic representation:} Word vectors such as Word2Vec and Glove are effective semantic representations of NLP tasks. We apply TextCNN \cite{kim-2014-convolutional} with Glove embedding as a benchmark. The Glove embedding vectors used in the experiments are trained on Wikipedia 2014 and Gigaword 5, with 300 dimensions. The TextCNN has kernel sizes of 3, 4, and 5 in the convolution modules.

\textbf{Word-level semantic representation with sequence modeling:} With word vectors, we further utilize the TextRCNN \cite{Lai_Xu_Liu_Zhao_2015} model to capture the sequential signal out of each individual word. TextRCNN and other RNN-based models utilize a bi-directional LSTM structure with hidden sizes of 200.


\textbf{Word-level sequence modeling with attention mechanism:} This model performs the best in predicting the MPAA ratings based on movie scripts and rich metadata \cite{shafaei-etal-2020-age}. For script text processing, they apply LSTM to model the sequential information from the word embeddings and use attention mechanism to aggregate the output of each time step for text representation.

\textbf{Pretrained language model task fine-tuning:} Contextualized representation from pretrained Transformer-based models have shown significant success on various NLP tasks. One popular variant, BERT \cite{devlin-etal-2019-bert}, was also used in classifying explicitness in the previous work\cite{fell2020love}. We fine-tune a pretrained BERT on this task in the multi-class classification setting. The BERT model is adapted from HuggingFace with a maximum input length limitation of 512 tokens.

\textbf{Sentence-level semantic representation with sequence modeling:} This model is the state-of-the-art in a severity rating problem for age-restricted content in movies \cite{zhang-etal-2021-none-severe}. We leverage the strong representation capability from the pretrained languages to obtain semantic representations. Then we apply general-purpose sentence embedding from Sentence-BERT \cite{reimers-gurevych-2019-sentence} to encode each sentence from the lyrics of a music item. Then the semantic representation sequences are further encoded using a recurrent architecture to model sequential information. It also becomes a part of the backbone model in our proposed method.

\textbf{Emotion-guided Transformer model:} We apply a Distilled RoBERTa model that is finetuned on emotion detection tasks \cite{hartmann2022emotionenglish} to obtain the emotion-guided sentence embeddings. 

\textbf{Multitask and ordinality-enforcement:} Our proposed multitask model predicts 5 aspects (\emph{Positive Messages}, \emph{Violence}, \emph{Substance Consumption} (Drinking, drugs, and smoking), \emph{Sex}, and \emph{Consumerism}) at one single prediction. The ordinality-enforcement components are applied to each individual aspect prediction.

All experiments are conducted using NVIDIA Tesla P40 and PyTorch 1.6.0/PyTorch Lightning 1.0.2. The optimizer is Adam optimizer with 0.001 as the learning rate. Each training epoch of the proposed method takes less than 30 seconds under a batch size of 40.
\subsection{Ranking-classification loss behavior}
\begin{figure*}[ht]
\centering
  \includegraphics[width=0.9\linewidth]{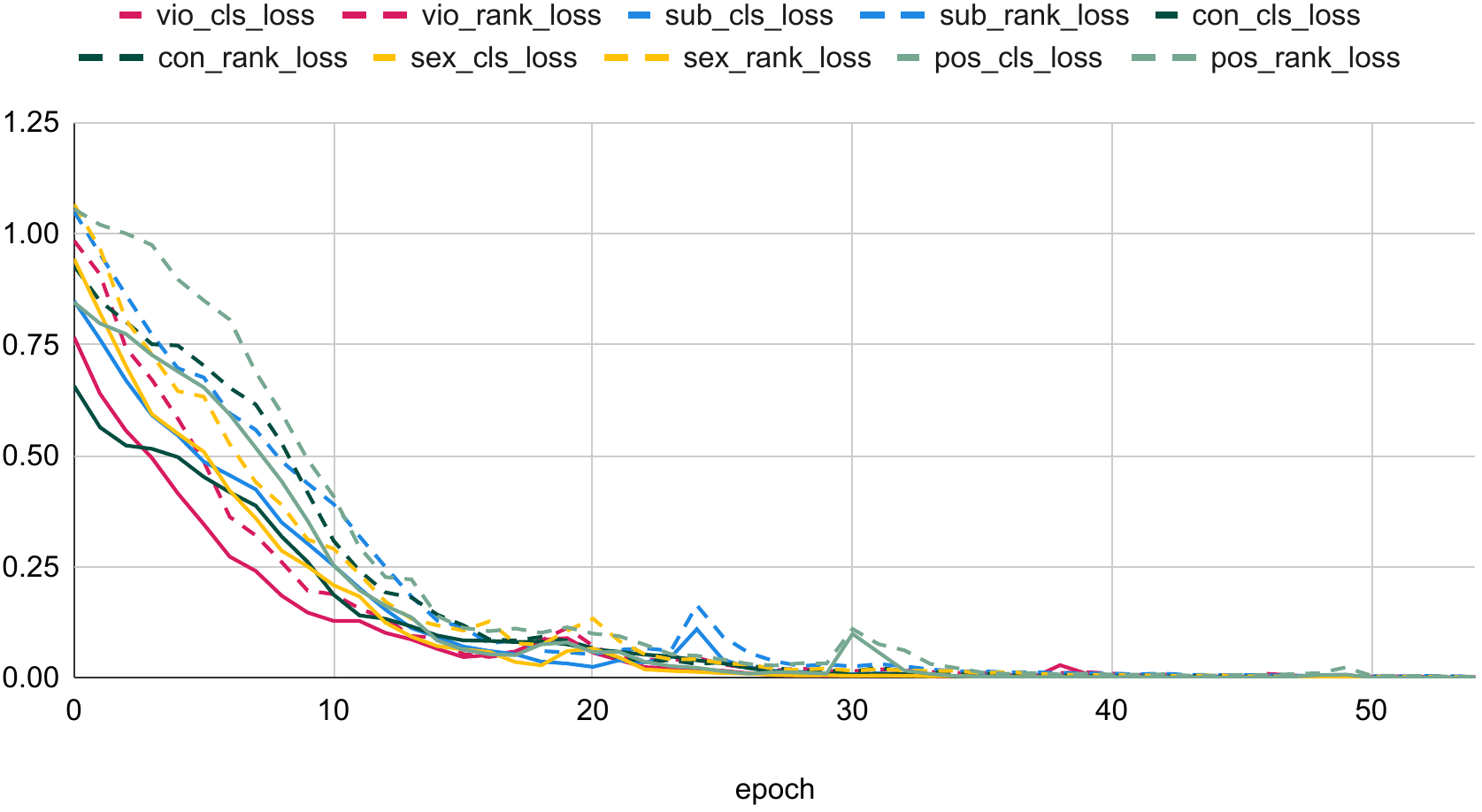}
  \caption{Training loss behavior of the ranking-classification method. The classification and ranking losses within one aspect are grouped by the same tone of the color palette: classification losses are in solid lines while ranking losses are in dash lines.}
  \label{fig:rankclsloss}
\end{figure*}

The training behavior of the ranking-classification joint loss in the ordinality-enforcement method is shown in Figure \ref{fig:rankclsloss}. Both cross-entropy losses are averaged on each loss instance in one batch. The ranking loss is often higher during the training process. We suppose ranking is more challenging because the ranking pairs are randomly constructed for every new training step.

\subsection{Prompt used for LLM evaluation}
\textbf{Simple context:} \texttt{Please assess the lyrics of the song given the following aspects: The lyrics: <Full Lyrics>
Positive Messages; 
Violence \& Scariness; 
Sex, Romance \& Nudity; 
Drinking, Drugs \& Smoking; 
Products \& Purchases (refer to Consumerism). \\
Please rate the presence of each aspect on a scale of 0 to 2, where 0 indicates 'low', 1 indicates 'medium', and 2 indicates 'high'. Provide your ratings strictly in the JSON format as shown in the example below and make sure no extra content: Example: \{"Positive Messages": 1, "Violence \& Scariness": 0, "Sex, Romance \& Nudity": 1, "Drinking, Drugs \& Smoking": 2, "Products \& Purchases": 1\}. }

\textbf{Rich context:} \texttt{Please assess the lyrics of the song given the following aspects: The lyrics: <Full Lyrics>
Positive Messages: <Full aspect description from CSM>; 
Violence \& Scariness: <Full aspect description from CSM>; 
Sex, Romance \& Nudity: <Full aspect description from CSM>; 
Drinking, Drugs \& Smoking: <Full aspect description from CSM>; 
Products \& Purchases (refer to Consumerism): <Full aspect description from CSM>. \\
Please rate the presence of each aspect on a scale of 0 to 2, where 0 indicates 'low', 1 indicates 'medium', and 2 indicates 'high'. Provide your ratings strictly in the JSON format as shown in the example below and make sure no extra content: Example: \{"Positive Messages": 1, "Violence \& Scariness": 0, "Sex, Romance \& Nudity": 1, "Drinking, Drugs \& Smoking": 2, "Products \& Purchases": 1\}. }

\textbf{Chain-of-Thought (CoT) prompt:}  \texttt{ \\(Keep context part the same). \\
Please rate the presence of each aspect on a scale of 0 to 2, where 0 indicates 'low', 1 indicates 'medium', and 2 indicates 'high'. Let's think step-by-step to analyze the lyrics and then provide your ratings in the JSON format. Here is an example output: This song promotes ... and the song has strong ... It depicts ..., so it implies ... (your chain-of-thought) ... Therefore, we reach the final assessment result: \{"Positive Messages": 1, "Violence \& Scariness": 0, "Sex, Romance \& Nudity": 1, "Drinking, Drugs \& Smoking": 2, "Products \& Purchases": 1\}. Now it is your turn: }

\end{document}